\documentclass[letterpaper, 10 pt, conference]{ieeeconf}  % Comment this line out if you need a4paper

\IEEEoverridecommandlockouts                              % This command is only needed if 
\usepackage{amsmath,amssymb,amsfonts}
\usepackage{algorithmic}
\usepackage{graphicx}
\usepackage{graphics}
\usepackage{textcomp}
\usepackage[table,xcdraw]{xcolor}
\usepackage{leftindex}
\usepackage{amsmath}
\usepackage{mathtools}
\usepackage{multirow}
\usepackage{booktabs}
\usepackage{svg}
\usepackage{hyperref}
\hypersetup{
    colorlinks=true,
    linkcolor=black,
    filecolor=magenta,      
    urlcolor=blue,
    }
\usepackage{calrsfs}
\usepackage{upgreek}
\usepackage{tikz}
\usetikzlibrary{3d,calc,intersections, positioning, shapes.geometric, fit}
\usepackage{mathrsfs} 
\usepackage{bm}
\usepackage{makecell}
\usepackage{arydshln}

\def\BibTeX{{\rm B\kern-.05em{\sc i\kern-.025em b}\kern-.08em
    T\kern-.1667em\lower.7ex\hbox{E}\kern-.125emX}}

\definecolor{blue_}{HTML}{3866D7}
\definecolor{orange_}{HTML}{EA8A60}
\definecolor{olive_}{HTML}{9B9536}  %former olive
\definecolor{cyan_}{HTML}{2AB1AD}
\definecolor{opt_flow}{HTML}{BA6560} %

\usepackage{etoolbox}
\makeatletter
\patchcmd{\@makecaption}
  {\scshape}
  {}
  {}
  {}
\makeatother

\newcommand{\topTableSpacing}{5pt}
\newcommand{\bottomTableSpacing}{3pt}

\title{\LARGE \bf
Object-Centric Action-Enhanced Representations for Robot Visuo-Motor Policy Learning
}

\author{
Nikos Giannakakis$^{1\dagger}$, 
Argyris Manetas$^{2\dagger}$, 
Panagiotis P. Filntisis$^{2}$,
Petros Maragos$^{3}$
and George Retsinas$^{2}$% <-this % stops a space
\thanks{$\dagger$ equal contribution}
\thanks{$^{1}$ N. Giannakakis is with 
 the School of ECE, National Technical University of Athens, Greece {\tt\footnotesize ngiannakakis@mail.ntua.gr}
 }
\thanks{$^{2}$ A. Manetas, P.P. Filntisis and G. Retsinas are with the Robotics Institute, Athena Research Center, 15125 Maroussi, Athens, Greece
{\tt\footnotesize a.manetas@athenarc.gr, pfilntisis@athenarc.gr,george.retsinas@athenarc.gr}
}
\thanks{$^{3}$
P. Maragos is with the School of ECE, NTUA, Athens, Greece, the Robotics Institute, Athena Research Center, Athens, Greece and the HERON - Hellenic Robotics Center of Excellence, Athens, Greece 
{\tt\footnotesize maragos@cs.ntua.gr}
}
}

\begin{document}

\maketitle
\thispagestyle{empty}
\pagestyle{empty}

% \begin{document}

% \title{VDO-LINE-SLAM*\\

% \thanks{Identify applicable funding agency here. If none, delete this.}
% }

% \author{\IEEEauthorblockN{Argyris Manetas}
% \IEEEauthorblockA{\textit{National Technical University of Athens}}
% \and
% \IEEEauthorblockN{Panagiotis Mermigkas}
% \IEEEauthorblockA{\textit{National Technical University of Athens}}
% \and
% \IEEEauthorblockN{Petros Maragos}
% \IEEEauthorblockA{\textit{National Technical University of Athens}}
% }

% \maketitle

\begin{abstract}
Learning visual representations from observing actions to benefit robot visuo-motor policy generation is a promising direction that closely resembles human cognitive function and perception. Motivated by this, and further inspired by psychological theories suggesting that humans process scenes in an object-based fashion, we propose an object-centric encoder that performs semantic segmentation and visual representation generation in a coupled manner, unlike other works, which treat these as separate processes. To achieve this, we leverage the Slot Attention mechanism and use the SOLV~\cite{SOLV} model, pretrained in large out-of-domain datasets, to bootstrap fine-tuning on human action video data. Through simulated robotic tasks, we demonstrate that visual representations can enhance reinforcement and imitation learning training, highlighting the effectiveness of our integrated approach for semantic segmentation and encoding. 
Furthermore, we show that exploiting models pretrained on out-of-domain datasets can benefit this process, and that fine-tuning on datasets depicting human actions ---although still out-of-domain---, can significantly improve performance due to close alignment with robotic tasks. These findings show the capability to reduce reliance on annotated or robot-specific action datasets and the potential to build on existing visual encoders to accelerate training and improve generalizability.
\end{abstract}

%\begin{IEEEkeywords}
%component, formatting, style, styling, insert
%\end{IEEEkeywords}

\section{Introduction}

In recent years, the intersection of machine learning and robotics have transformed our understanding of how machines perceive and interact with their environments. This has encouraged researchers to explore more deeply the underlying principles of human learning, motivation and action, delving into the field of psychology, to often guide their algorithm or system design. One such area investigates how different individuals perceive objects and other parts of the scene to construct internal representations that capture feasible actions on them ---these action possibilities constitute what is referred to in literature as affordances. Specifically, according to J.J. Gibson~\cite{gibson_affordances}, who termed this concept, objects have inherent ``values" or ``meanings"~\cite{affordances_donc}, that observers uncover through interaction and exploration, linking them to potential actions.
Children, for example, perceive the shapes, sizes, colors, motions, and other characteristics of objects and are able to learn how things can be manipulated, e.g. moved, squeezed, stacked, and others. Although all senses participate in this process, the question of how artificial intelligence can construct \textit{visual} representations to mimic the way humans perceive such cues, has attracted considerable attention, due to the widespread availability of cameras.\par
%TODO: is dynamo used for visuo-motor pretraining?
\begin{figure}[t] % or [h] or [!htbp] depending on placement
  \centering
  \includegraphics[width=\columnwidth]{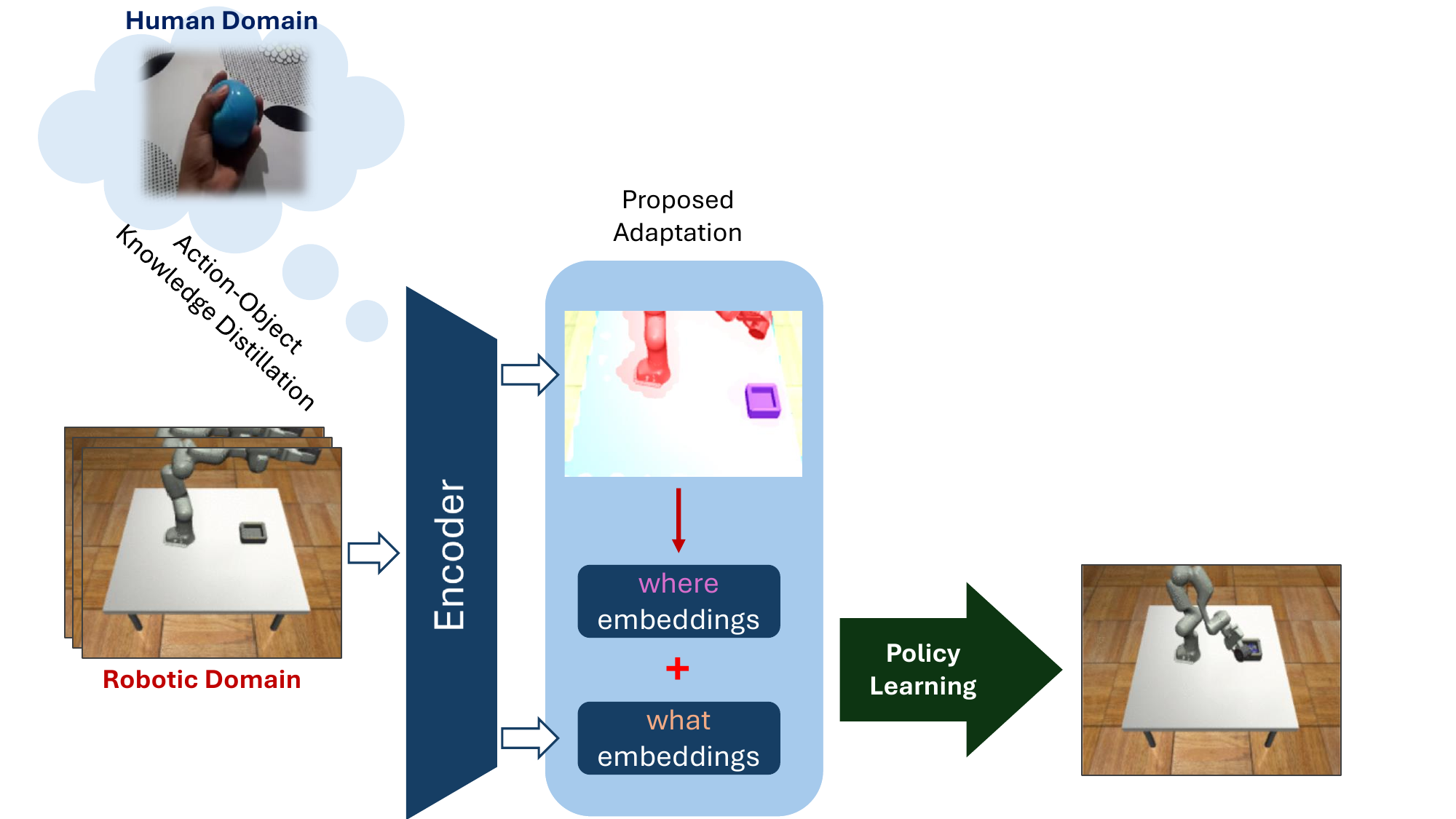}
  \caption{\textbf{Approach Overview:} Our proposed encoder extracts ``what" and ``where" visual object-centric embeddings, which are combined to form scene representations. These embeddings distill action-based knowledge acquired from pretraining on human action videos to effectively guide robot policy learning.}
  \label{fig:your_label}
\end{figure}
These visual features can complement explicit state representations, enhancing visuo-motor policy training in reinforcement and imitation learning scenarios, as demonstrated in previous works~\cite{vip, dynamo}. Approaches following this paradigm typically pre-train neural networks in an unsupervised manner ---using either in-domain or out-of-domain datasets, such as~\cite{ego4d, Somethingsomething, imagenet, robonet}--- before applying them to robotics tasks. This strategy has been shown to reduce training time and improve performance and generalization compared to end-to-end learning methods~\cite{deep_l_for_robot_per,learn_navig_mid, toto}. The resulting learned representations can subsequently be fine-tuned or directly employed for downstream robot manipulation tasks. Particularly when leveraging video datasets, the sequential structure naturally distills embeddings with information about \textit{potential} dynamic behavior of parts of the scene ---objects and others---, closely resembling human cognitive development through everyday interaction with the world. For example, infants intuitively lift, push and drop objects, observing the outcomes and incrementally constructing their world knowledge.\par

\begin{figure*}[t]
    \centering
    \includegraphics[width=0.7\textwidth]{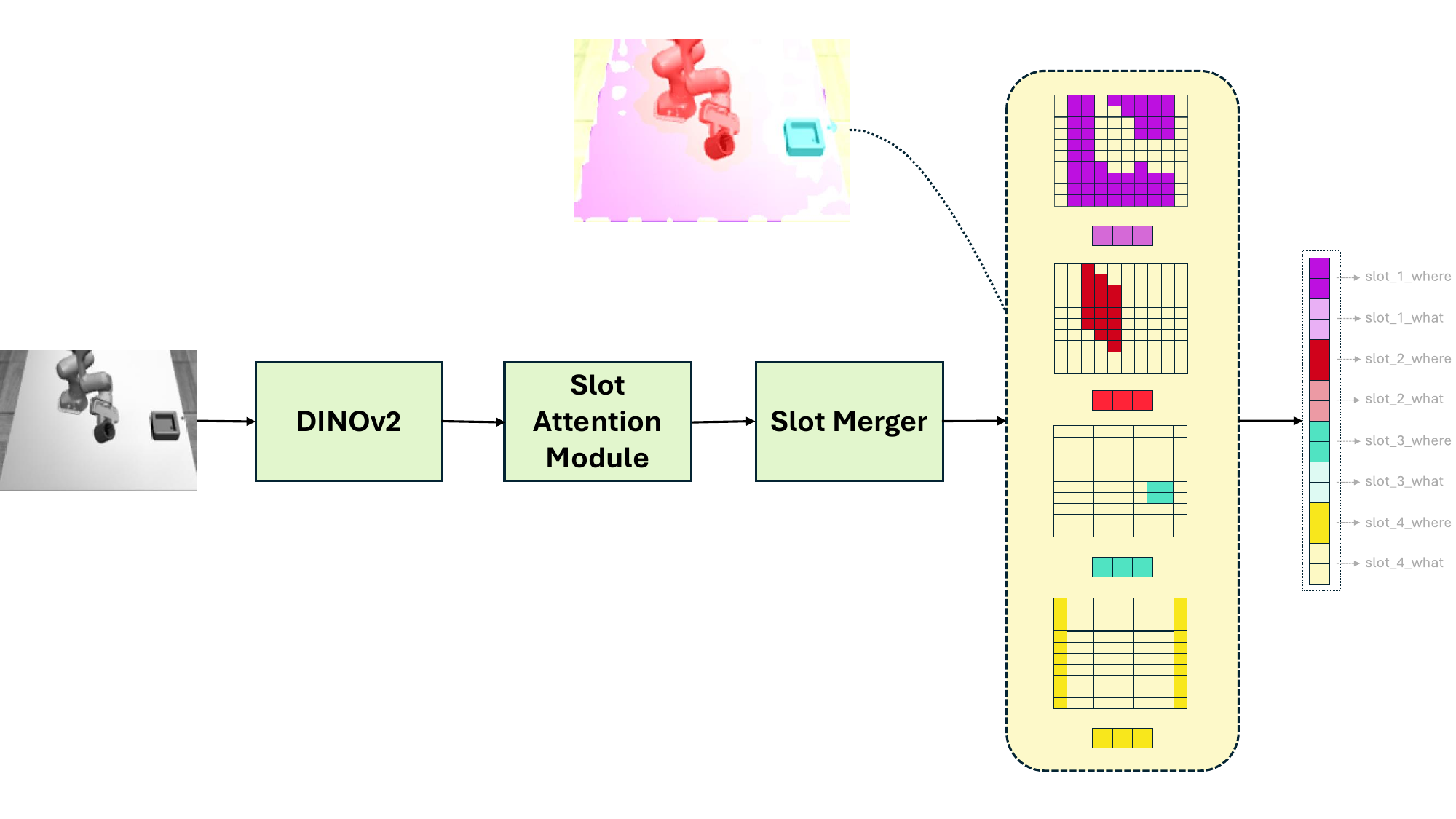}
    \caption{\textbf{Our proposed architecture:} Image embeddings produced by DINOv2 are processed by the Slot Attention Module to generate slots. The Slot Merger then outputs four final object-specific features, referred to as ``what" vectors, each corresponding to a distinct semantic region. These are combined with the associated ``where" vectors, which encode spatial information derived from the attention masks, to form a unified object-centric scene representation.}
    \label{fig:encoder}
\end{figure*}

In line with these observations, an emerging direction in machine learning and computer vision is object-centric representation learning, which aims to represent complex environments in terms of discrete objects, rather than treating an entire scene as a single entity. This approach is backed by and aligns closely with the proposed ``principles of grouping" from psychology and cognitive science, which describe how humans process visual signals by organizing them into distinct objects~\cite{principles_grouping}. Consistent with this, experiments in developmental psychology show that infants learn very early on action-object associations; even before word associations, which become more important later in development~\cite{word_associations}. Systems following this object-centric paradigm have shown promising results~\cite{viola, pocr}, while also offering improved interpretability. This underscores the significance of incorporating these human development insights into the design of representation encoding methods.\par

Nevertheless, many object-centric methods decouple the problems of object segmentation and visual representation generation, potentially introducing propagating errors ---in the case of inaccurate semantic segmentation--- and requiring that all visible object classes in the frames be known in advance. To overcome these limitations, we propose an encoder capable of producing object-centric representations that capture information about both the characteristics of the objects and their position. Our system leverages the SOLV~\cite{SOLV} framework, which builds on the slot attention mechanism of previous works~\cite{slot_attention, inv_slot_Attention} to enable the concurrent object scene segmentation and object-specific vector representation generation. 
Capitalizing on the pretrained nature of SOLV, we fine-tune it on a subset of the Something-Something action video dataset~\cite{Somethingsomething} to develop an encoder that outputs object-centric visual features, which are aggregated into a common scene representation encompassing potential actions and spatiality information of the visible objects. We evaluate the effectiveness of this object-centric scene representation to guide robot visuo-motor policy generation in reinforcement and imitation learning scenarios, comparing them against state-of-the-art encoders, demonstrating that our encoder, benefits from the object-centric approach, and that knowledge acquired from human action videos transfers well to robotic tasks, significantly enhancing the robot's efficacy.\par
Our key contributions can therefore be summarized as follows:\begin{itemize}
    \item We develop an object-centric visual encoder by adapting the SOLV framework for visuo-motor policy learning.
    \item We fine-tune the encoder on videos depicting human actions to distill action-relevant knowledge into the representations.
    \item We evaluate the effectiveness of these representations on imitation and reinforcement learning scenarios, outperforming state-of-the-art image encoders on the TOTO~\cite{toto} benchmark.
\end{itemize}

%is how humans create visual representations of scenes they are seeing, encompassing into these information about possible actions that can be done on them, a term that is referred in the bibliography as affordances, and how could Artificial Intelligence possibly mimic that. 
%It is a concept backed both intuitevely 
\section{Proposed Method}
\label{sec:proposed_method}

\begin{figure*}[t]
    \centering
    \includegraphics[width=0.3\textwidth]{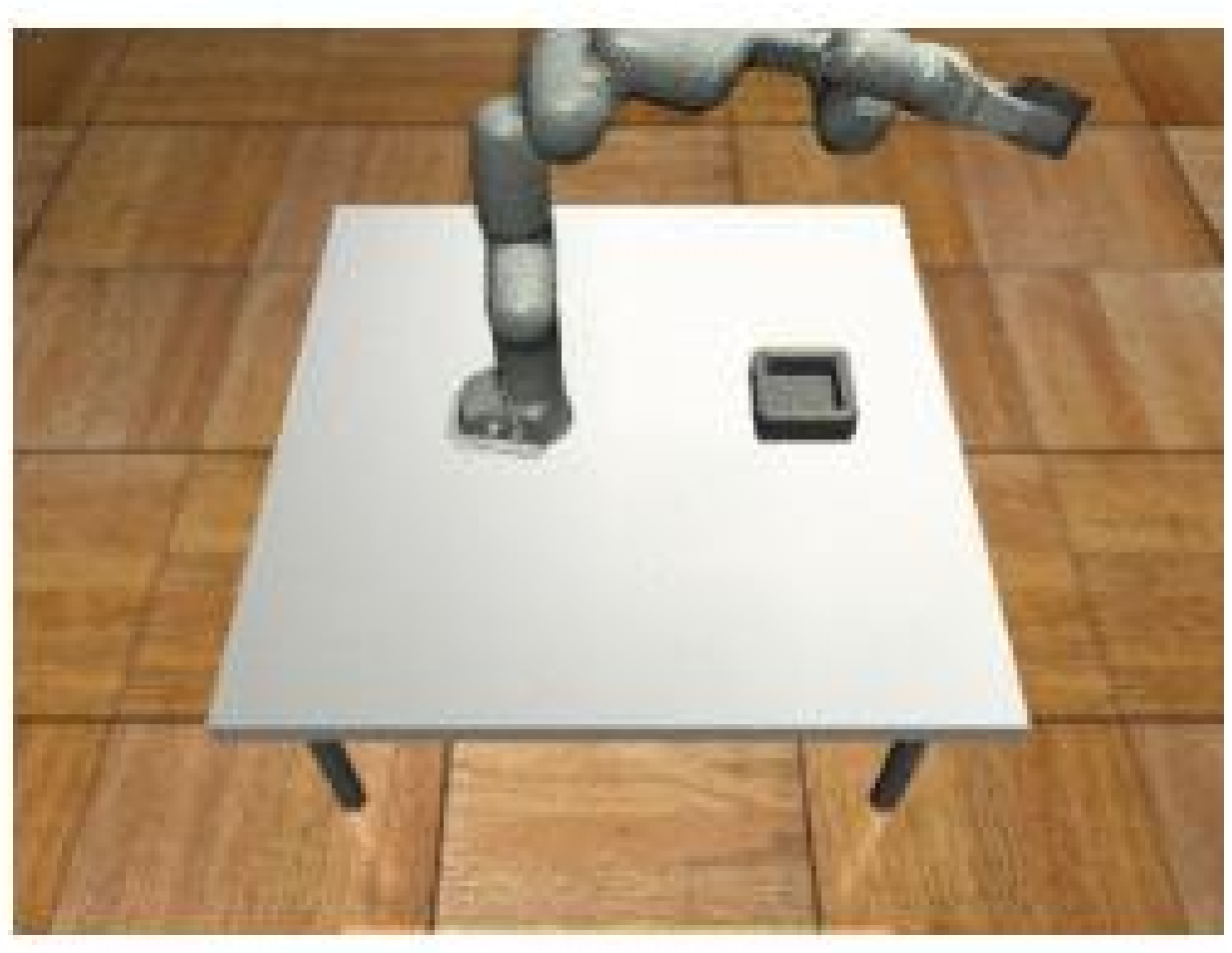}
    \includegraphics[width=0.3\textwidth]{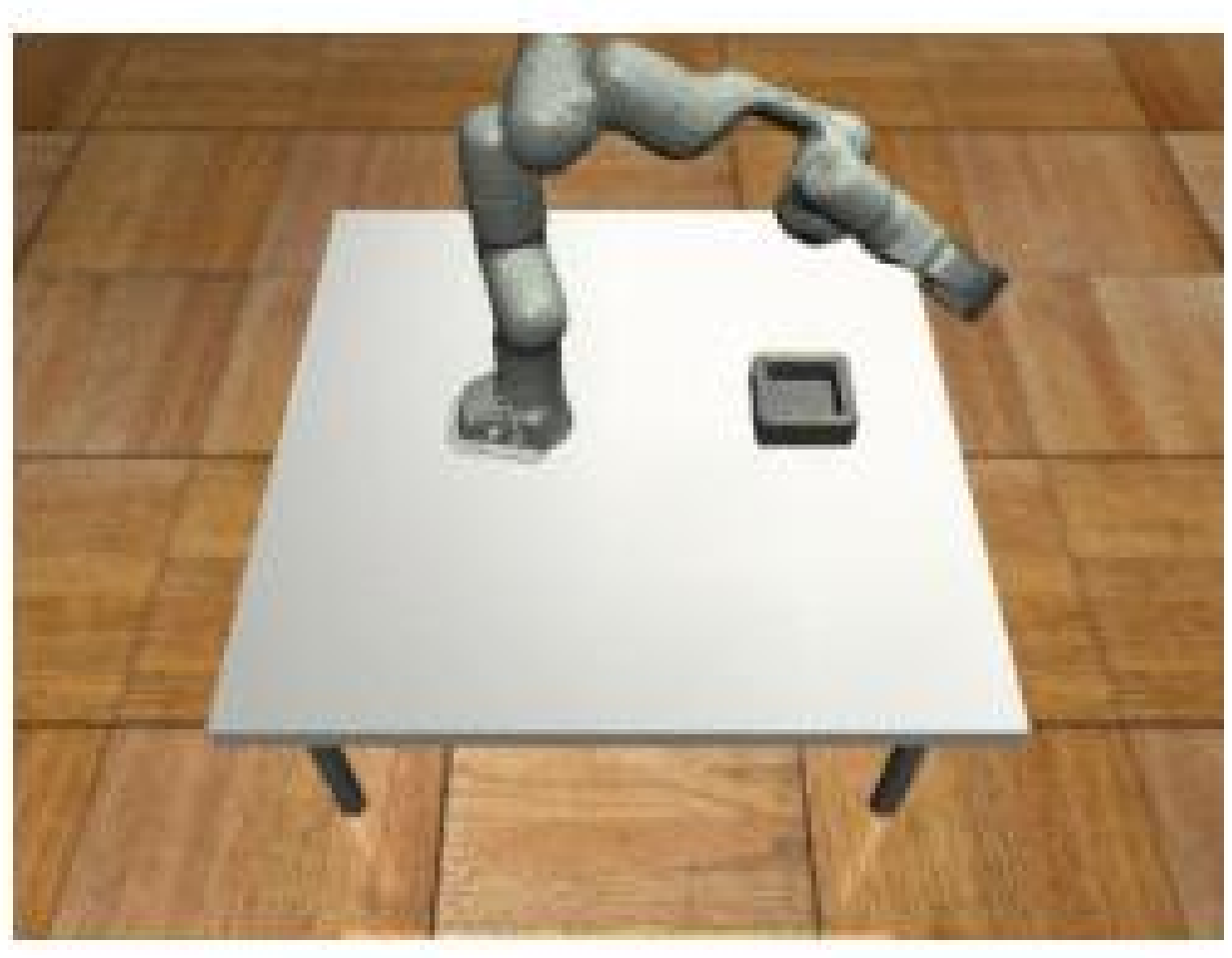}
    \includegraphics[width=0.3\textwidth]{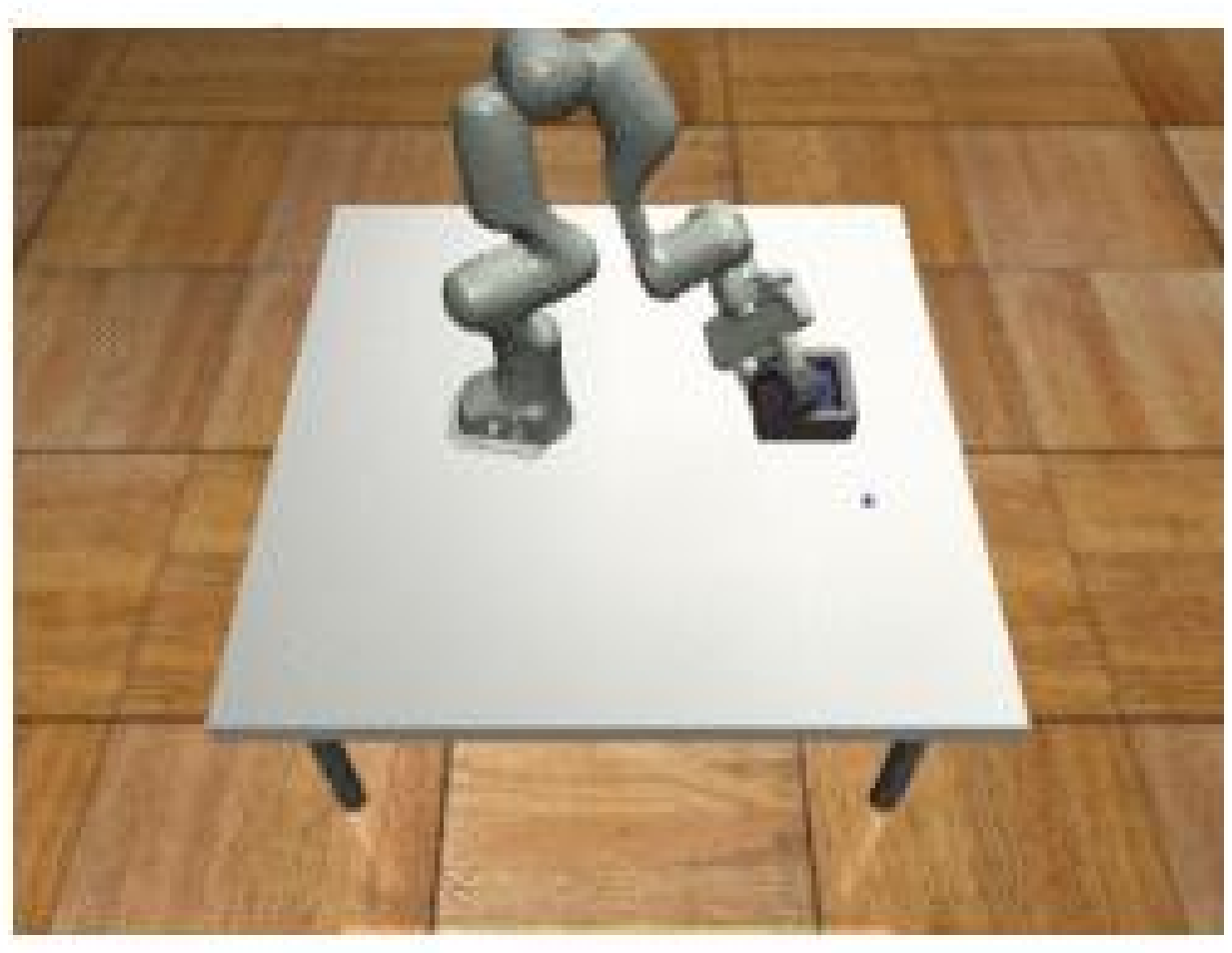}
    \caption{\textbf{Experimental Environment:} We simulate the Franka Emika Panda robot arm pouring task within the TOTO framework. The goal is to transfer as many spheres as possible from the cup to the container.}
    \label{fig:toto_benchmark}
\end{figure*}

An outline of our proposed method can be seen in Figure~\ref{fig:encoder}. Our proposed methodology is designed to be domain- and task-agnostic, allowing the encoding of object- and scene-specific information into representations that can be utilized in reinforcement and imitation learning scenarios for relevant robotic task completion. Given our object-centric focus, our work builds on recent advances in slot attention mechanisms and SOLV (Self-Supervised Object-Centric Learning for Videos), a self-supervised framework capable of segmenting and tracking multiple objects in real-world video sequences, without the need for cumbersome annotations or additional modalities such as depth and optical flow. In the following paragraphs, we present the core ideas that shaped our approach and the main components of our architecture. \par
A key component of our architecture is the slot attention module, introduced in~\cite{slot_attention} and later extended in~\cite{inv_slot_Attention} to enforce transformation-invariant visual representations. This mechanism assigns regions of an image or video frame to a set of ``slots", which are learnable vectors that represent distinct objects or components in the scene. Each slot is updated iteratively by attending to the input features, allowing it to bind to a specific object or semantic region, thus producing object-centric representations without supervision.\par

SOLV~\cite{SOLV} leverages the slot attention mechanism introduced in~\cite{inv_slot_Attention} to produce spatial slots that are associated across video frames, resulting in temporally-aware object representations. This makes SOLV a natural choice for our setting, where understanding potential object behavior is essential. Specifically, by fine-tuning on videos depicting human actions, we aim to encourage the model to capture action-relevant features, such as possible object dynamics and affordances, thereby making the resulting representations more effective in robot policy learning.

%In our approach, specifically, we adopt the variant proposed in~\cite{inv_slot_Attention}, which creates consistent object representations, regardless of scale and translation.\par

% Mention that even though the invariant slots are used, there is some temporal encoding?
%Building upon this foundation, we slightly augment the definition of slots and modify the SOLV architecture
Accordingly, we slightly augment the definition of slots and modify the SOLV architecture to create an encoder that generates suitable object-centric representations for visuo-motor policy learning  following the same principles as in~\cite{viola, pocr}, for every video frame. 
Our encoder implementation consists of the following SOLV components\footnote{for more details on the SOLV components and their implementation see the original paper~\cite{SOLV}}:
\begin{itemize}
    \item \textbf{Visual Backbone}: The frozen DINOv2~\cite{dinov2} encoder extracts visual embeddings from the input video frames.
    \item \textbf{Slot Attention}: The slot attention mechanism binds slots into semantically different regions for each frame, following the paradigm of~\cite{inv_slot_Attention}, resulting in object segmentation. 
    Shared initialization is used across time steps to ensure temporal consistency.
    \item \textbf{Slot Merging}: This module merges slots based on their similarity, through a configured Agglomerative Clustering (AC) algorithm, which outputs 4 final slots ---a quantity which through our experimentation ensured the best mask granularity avoiding over- and under-segmentation (see Section~\ref{sec:num_slots}).
\end{itemize}

Unlike SOLV, we do not include a temporary binding step in the final encoder (referred to as the \textit{Temporal Binding Component} in the original SOLV paper), as we argue that the slot vectors produced by the slot attention module already capture non-trivial temporal information, akin to a video-to-image distillation process, and thus some action-object affordance association for the target robotic tasks. That is, because the slot attention module is pretrained in a system that processes temporal information, the resulting slots are optimized for inter-frame attention, and hence inherently capture temporal signals.

%Unlike SOLV, we do not include a temporary binding step in the final encoder (referred to as the \textit{Temporal Binding Component} in the original SOLV paper), as we argue that the \textcolor{red}{spatial} slot vectors produced by the spatial binding component already capture non-trivial temporal information, akin to a video-to-image distillation process, and thus some action-object affordance association, for the target robotic tasks. 
%Since, the spatial binder is pretrained in a system that processes temporal information, these slots are optimized for inter-frame attention, and hence inherently capture temporal signals. \par

We augment the final slots, by introducing an explicit spatial component, following~\cite{pocr}. The initial four slots, produced by iterative slot binding, with a dimension of $D_{\text{what}} = 128$, encode objects' intrinsic feature cues, forming what we term the ``what" component. Since these representations are to be used to generate the visuo-motor policy of robots, adding explicit information about the objects' position is intuitive. We achieve this by processing the attention masks of each slot, which are resized to dimensions $H'_{\text{att}} \times W'_{\text{att}} = 10 \times 10$ from $H_{\text{att}} \times W_{\text{att}} = 24 \times 36$ via bilinear interpolation. The scaled softmax activation function is applied to the resized attention masks, amplifying the regions that already had the highest values ---an adjustment that has demonstrated improved performance in our experiments. The resulting attention masks are flattened to produce the localization information vectors, hereinafter referred to as the ``where" vectors, which in combination with the ``what" vectors produce the final object-centric representation of the image, with a shape of $(D_{\text{what}} + D_{\text{where}}) \times \text{``number of slots"} = (128 + 100) \times 4 = 912$.

\section{Experimental Evaluation}
To study the ability of our encoder's object-centric representations in guiding a robotic learning task, we incorporate them into reinforcement and imitation learning scenarios as part of the state space, and compare their effectiveness against representations from other state-of-the-art image encoders that lack action-object association knowledge. 

\subsection{Pretraining and Fine-Tuning}
In our approach, we leverage SOLV’s ability to generate object-centric representations, acquired through training on large-scale datasets for semantic segmentation, and adapt it to capture action-object associations by fine-tuning its pretrained weights on video datasets containing such interactions. Specifically, to improve its capacity for constructing object-centric representations relevant to robot-specific tasks ---such as rolling, squeezing, and other interactions--- we fine-tune it for 100 epochs on a subset of the Something-Something~\cite{Somethingsomething} dataset, which includes such actions. The resulting network layers are then integrated into the final encoder described in Section~\ref{sec:proposed_method}. %TODO: Να προσθέσω ότι we show in the following experiments, that the videos of showing actions done on objects impact positevely the results even though they are done by humans, showing that they capture some intrinsic dynamic capabilities of the objects.

\subsection{Experimental Environment}

We assess the capabilities of our approach on the Train Offline, Test Online (TOTO)~\cite{toto} benchmark, which provides a protocol for evaluating both visual representations and policy learning. TOTO is a robotics benchmark designed to address the lack of standardization across research centers, providing remote research teams with access to shared robotic hardware and an open-source dataset of tasks for offline training. This dataset consists of noise-augmented trajectories collected through robot teleoperation, and trajectories generated by Behavior Cloning (BC) trained agents. The benchmark focuses on two manipulation tasks: (i) pouring and (ii) scooping, which despite being commonplace for humans, pose significant challenges to robots, due to variations in initial conditions and the objects involved, thus offering valuable insights into the methods under evaluation.\par
The TOTO benchmark includes a simulation package, which uses the MuJoCo physics simulation engine~\cite{mujoco}, and simulates the 7 degrees-of-freedom Franka Emika Panda robot arm~\cite{franka_emika}. Joints are restricted within a specified range of positions, offering the two-fold advantage of a smaller control space and improved safety in real-world deployments.\par
In our test setup, the simulated robot arm is initialized in random positions, holding a cup filled with twelve small spheres, with the goal of pouring as many of them into another container. A successful manipulation corresponds to experiments where at least one sphere is deposited in the container, while the reward is the percentage of the total spheres being successfully deposited. The training set used consists of 82 successful and 21 failed trajectories. In the following subsections, we present the results obtained from our imitation and reinforcement learning experiments.

\subsection{Imitation Learning}
\label{subsec:imitation_l}
In our first approach, we employ Imitation Learning (IL) to train the robot's policy, using expert demonstrations. Unlike reinforcement learning, where an agent learns by exploring and interacting with the environment, IL attempts to replicate expert behavior, avoiding the high costs, risks, and time demands of real-world exploration.
Among IL methods, we adopt Behavioral Cloning (BC)~\cite{behavior_cloning}, one of the earliest and simplest approaches. BC formulates policy learning as a supervised learning problem, where the agent directly maps states to expert actions without requiring exploration.\par
We train separate BC agents for each encoder under evaluation ---our own, and those we compare against--- using the 82 successful out of the 103 total trajectories (i.e., those with a reward above zero), with inputs produced by the corresponding encoder. During training the BC algorithm trains a neural network to serve as the agent's policy, receiving as input the encoders' result concatenated with the robot's current joint angles. The network predicts joint angle targets, which are fed to the MuJoCo controller to move the simulated robot arm. Therefore, the input dimension is $R_{\text{dim}} + 7$, where $R_{\text{dim}}$ is the size of the representation vector, while the output dimension is $\text{out}_{\text{dim}} = 7 \times h$, where $h$ (in our experiments $h=10$) represents the prediction horizon.\par
Our encoder, A-SOLV (AE), is compared against state-of-the-art image encoders: BYOL~\cite{BYOL}, CLIP~\cite{CLIP}, DINOv2~\cite{dinov2}, MoCo~\cite{MoCo}, and ResNet50~\cite{resnet}. 
Additionally, we evaluate our adapted encoder architecture using the pretrained weights from SOLV (referred to as A-SOLV), trained on the Youtube-VIS 2019~\cite{youtube_vis} real-world video dataset, to assess the impact of action-object associations introduced through fine-tuning on action videos from a subset of the Something-Something dataset. Each BC agent is trained for 80 epochs with a 0.001 learning rate. 
For each image encoder, we train 5 BC agents, and evaluate them on 100 randomly initialized trajectories to reduce stochasticity and obtain more reliable performance estimates. The results are summarized in Table~\ref{table:imitation_results}, showing that our encoder consistently achieves the highest reward and success rate.

\begin{table}[t]

\setlength{\dashlinedash}{4pt}
\setlength{\tabcolsep}{5pt}
\newlength\yexp
\settowidth{\yexp}{0.410}
\centering
\caption{\small Comparison of Pre-trained Visual Representation Models in Training Behavior Cloning Agents for the TOTO Benchmark Simulated Pouring Task. *Adapted SOLV, **Adapted SOLV (Action Enhanced).} %For the SOLV model, the representation size consists of 4 slots with representations of size 128, along with the corresponding scaled-down attention of size 100 each (total size 912).}
\resizebox{\columnwidth}{!}{%
\begin{tabular}{llccc}
\toprule[1.5pt] \addlinespace[\topTableSpacing]
Encoder & Dataset & Repr. Size & Success Rate & Mean Reward \\ \midrule
BYOL & ImageNet~\cite{imagenet} & 512 & 0.46 ± 0.05 & 17.48 ± 2.40 \\
CLIP & WIT (WebImageText)~\cite{CLIP} & 512 & 0.49 ± 0.06 & 18.61 ± 4.09 \\
DINOv2 & LVD-142M~\cite{dinov2} & 768 & 0.55 ± 0.04 & 18.72 ± 1.78 \\
MoCo & ImageNet~\cite{imagenet} & 2048 & 0.31 ± 0.04 & 9.40 ± 2.16 \\
ResNet50 & ImageNet~\cite{imagenet} & 2048 & 0.56 ± 0.11 & 21.15 ± 5.60 \\[0.15cm]
\cdashline{1-5}\noalign{\vskip 0.15cm}
A-SOLV* & Youtube-VIS 2019~\cite{youtube_vis} & 912 & 0.46 ± 0.06 & 15.07 ± 2.24 \\
A-SOLV (AE)** & Something-Something~\cite{Somethingsomething} & 912 & \textbf{0.61 ± 0.04} & \textbf{25.25 ± 2.95}\\
\addlinespace[\bottomTableSpacing] \bottomrule[1.5pt]
\end{tabular}
}
\label{table:imitation_results}
\end{table}

\subsection{Implicit Q-Learning}
\begin{figure*}[t]
    \centering
    \includegraphics[width=0.32\textwidth]{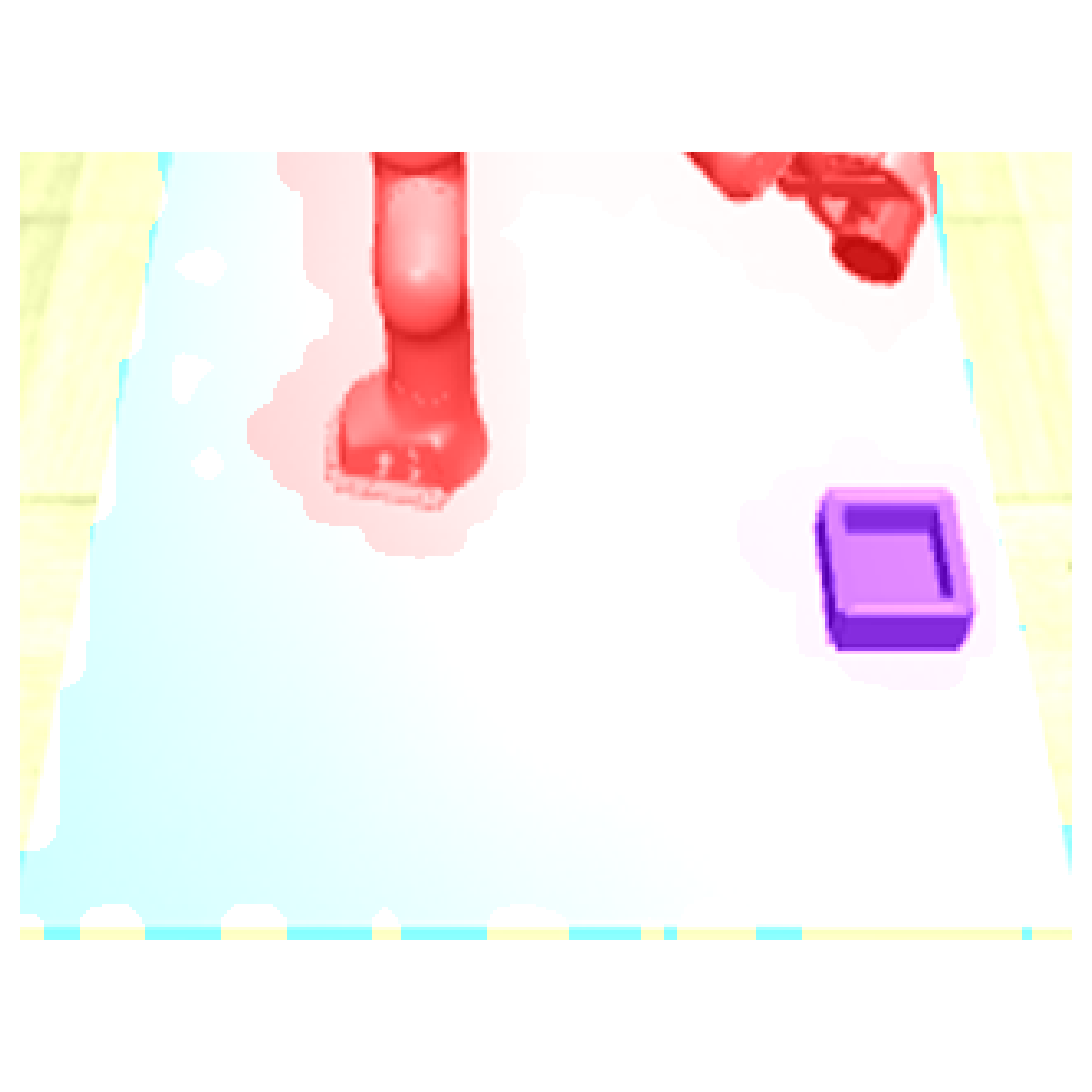}
    \hfill
    \includegraphics[width=0.32\textwidth]{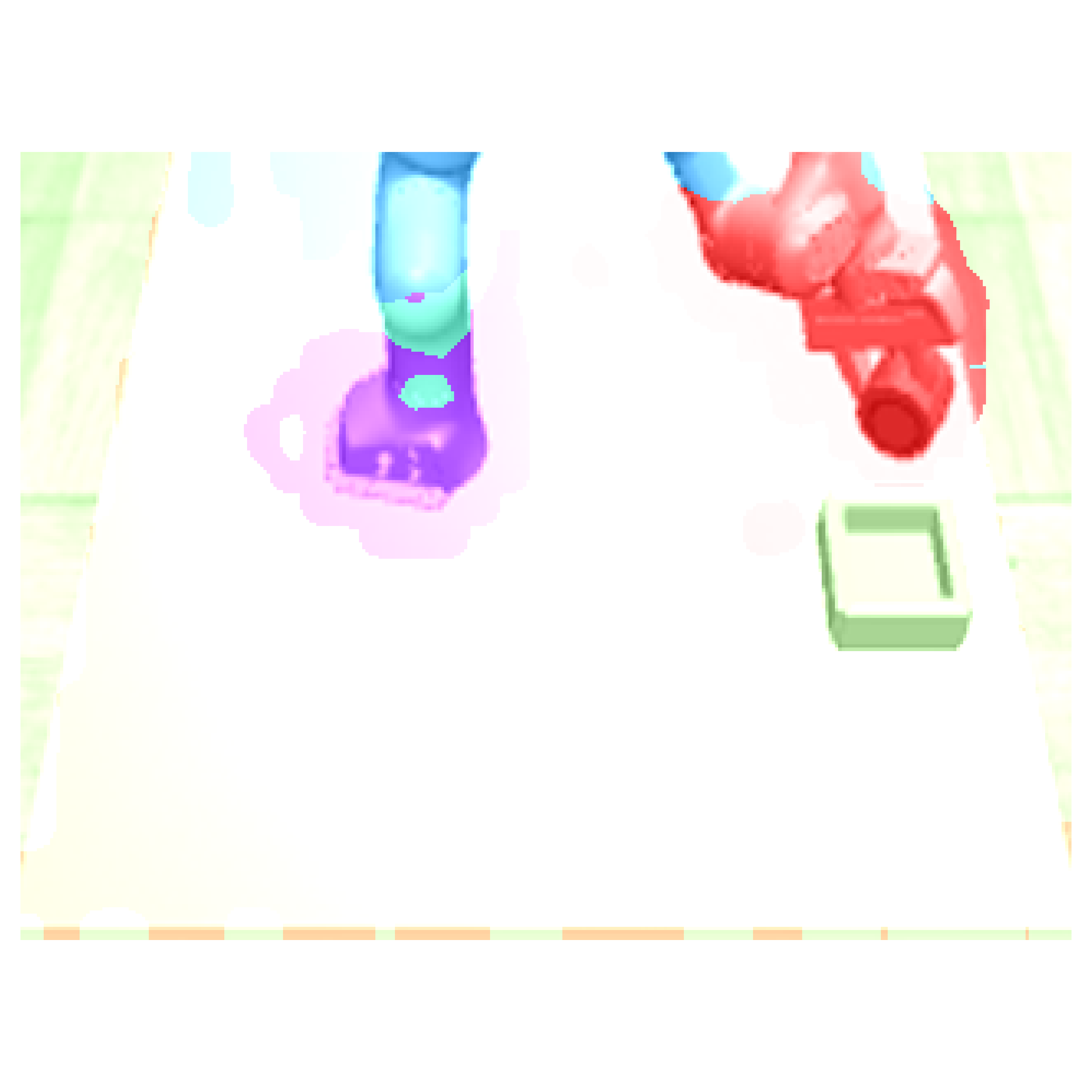}
    \hfill
    \includegraphics[width=0.32\textwidth]{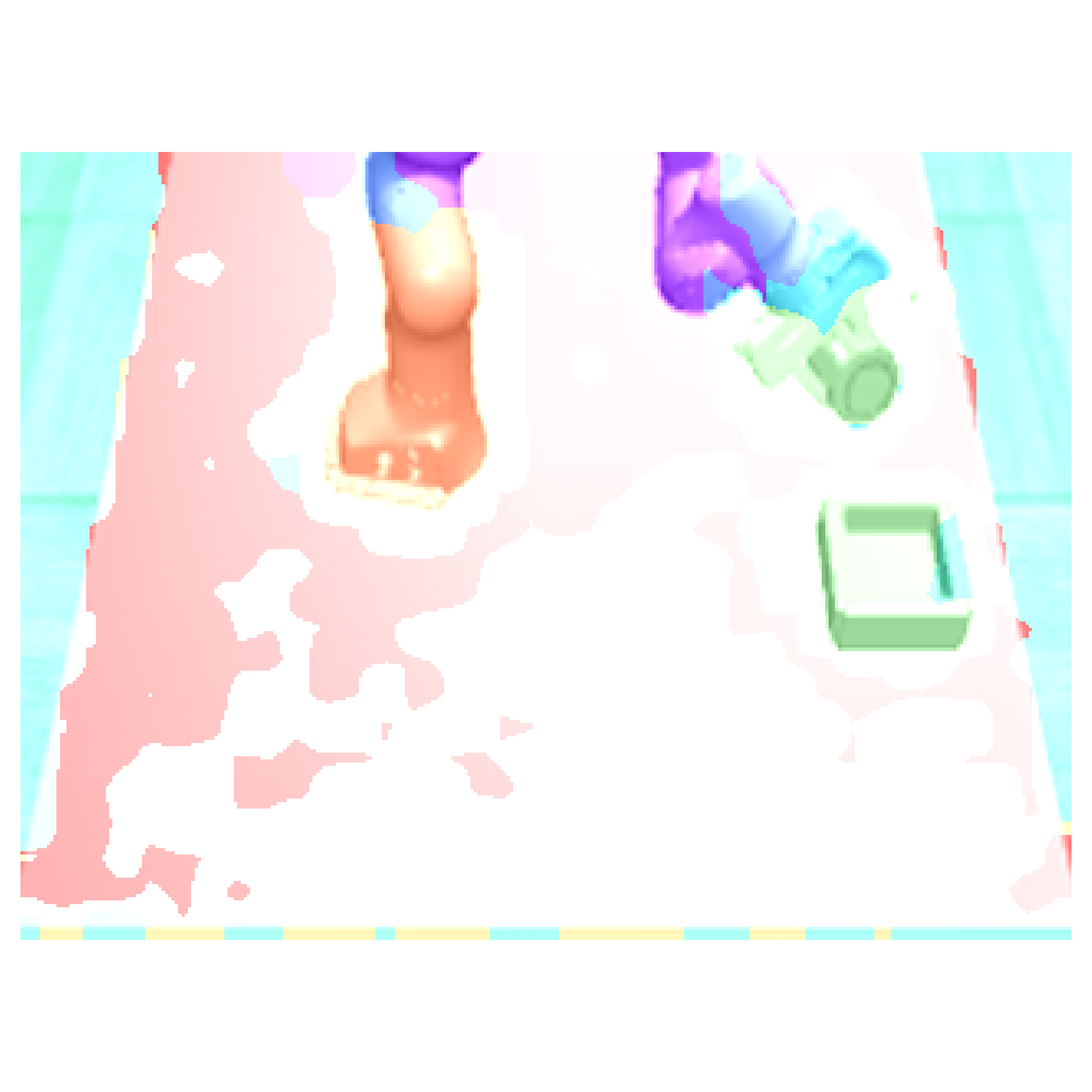}

    \caption{\textbf{Simulation frames from the TOTO pouring task, segmented by the Slot Attention masks of the SOLV model,} with the Slot Merger module outputting 4, 6, and 8 slots respectively (from left to right).}
    \label{fig:toto_slots}
\end{figure*}

\begin{figure*}[t]
    \centering
    \includegraphics[width=0.48\textwidth]{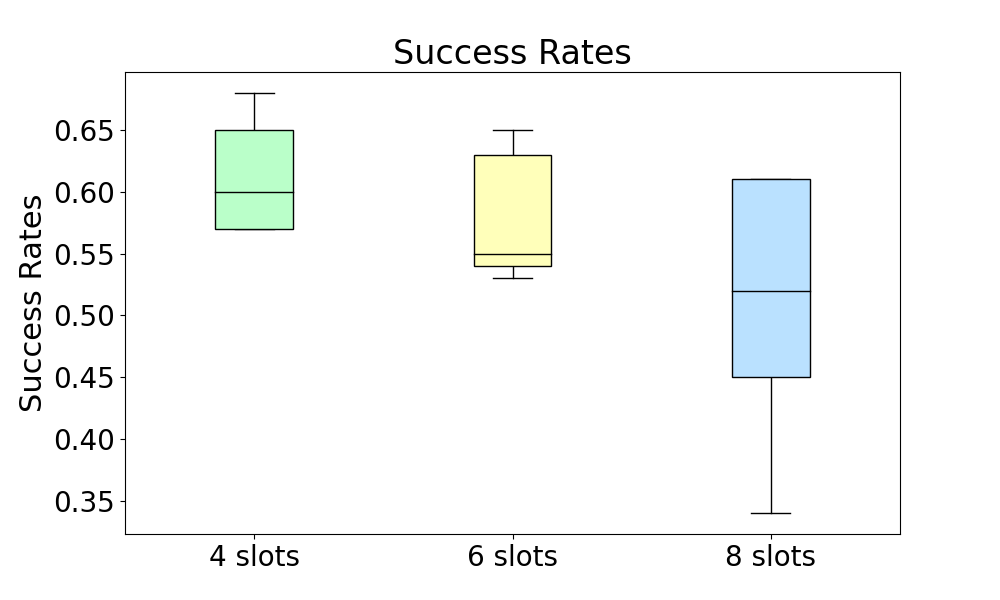}
    \hfill
    \includegraphics[width=0.48\textwidth]{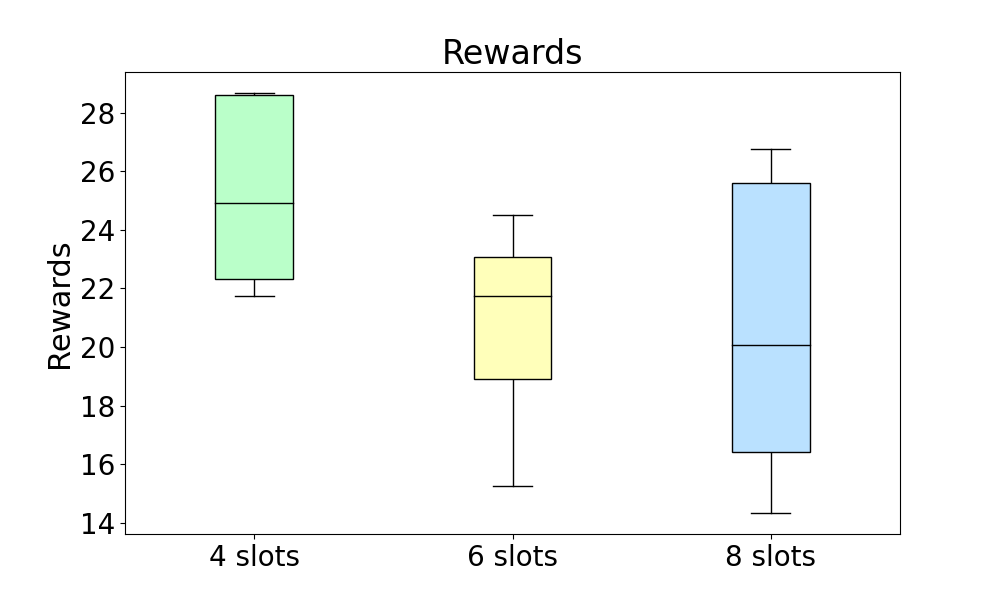}
    \caption{Success rates and mean rewards of different number of slots configurations in the TOTO pouring simulation task. Each variant was trained five times and evaluated across 100 trajectories.}
    \label{fig:slots}
\end{figure*}

\begin{table}[t]
\setlength{\tabcolsep}{5pt}
\centering
\caption{\small Encoder Comparison in the Implicit Q-Learning Scenario. *Adapted SOLV, **Adapted SOLV (Action Enhanced).}
\begin{tabular}{lccc}
\toprule[1.5pt]
Encoder & Representation Size & Success Rate & Mean Reward \\ \midrule
BYOL & 512 & 0.34 & 10.83 \\
ResNet50 & 2048 & 0.46 & 14.5 \\[0.15cm]
\cdashline{1-4}\noalign{\vskip 0.15cm}
A-SOLV* & 912 & 0.43 & 18.0 \\
A-SOLV (AE)** & 912 & \textbf{0.57} & \textbf{24.41}\\
\bottomrule[1.5pt]
\end{tabular}
\label{table:implicit_q_learning}
\end{table}

The Imitation Learning experiments in subsection~\ref{subsec:imitation_l} demonstrated the effectiveness of object-centric representations in improving robot policies learned from expert demonstrations, but to further evaluate their impact in standard reinforcement learning algorithms, we conduct experiments using Implicit Q-Learning (IQL)~\cite{implicit_q_learning}, an offline reinforcement learning algorithm that is designed to improve stability and performance by avoiding out-of-distribution Q-value queries. IQL achieves this by employing expectile regression to approximate optimal Q-values, mitigating extrapolation errors from unseen actions present in previous methods, delivering state-of-the-art results, thus rendering this approach particularly well-suited for our offline scenario.\par
This time, our training dataset consists of all 103 TOTO trajectories, which we use to train an IQL agent. We compare the effectiveness of our encoder against BYOL, ResNet50, and A-SOLV. The results are presented in Table~\ref{table:implicit_q_learning}. 

\subsection{Number of Slots}
\label{sec:num_slots}

To determine the optimal number of slots output by the slot merger module for achieving the best segmentation resolution ---with the least over- or under-segmentation of objects--- we conducted experiments with the AC algorithm configured to produce 4, 6, and 8 slots. Based on qualitative analysis of the resulting segmentations (see Figure~\ref{fig:toto_slots}) and a re-evaluation of the imitation learning experiments (see Figure~\ref{fig:slots}), we found that using 4 slots yielded the best results. 

\subsection{Contribution of the ``where" Components}
\label{sec:where_contr}

\begin{figure*}[t]
    \centering
    \includegraphics[width=0.48\textwidth]{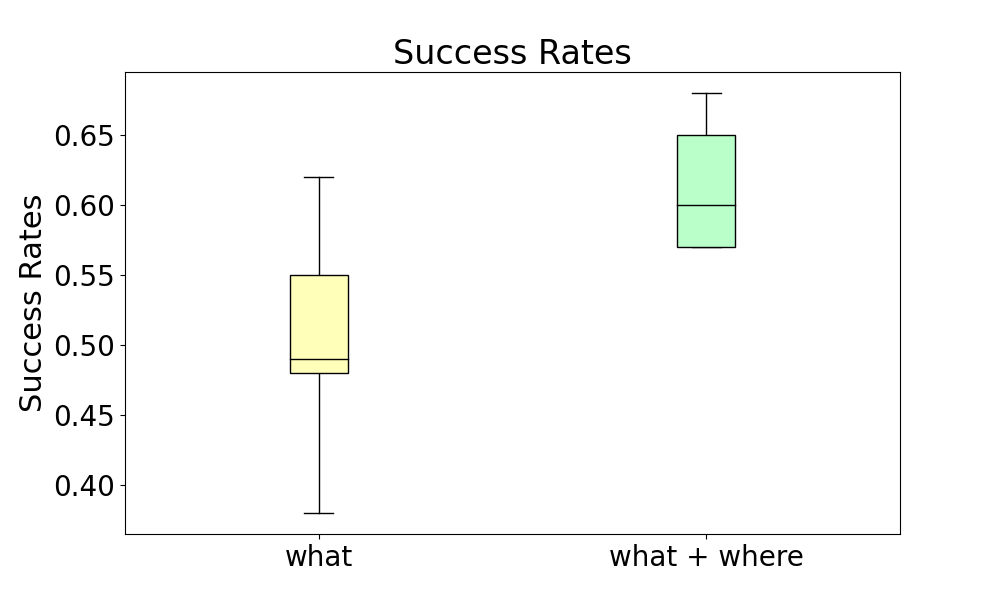}
    \hfill
    \includegraphics[width=0.48\textwidth]{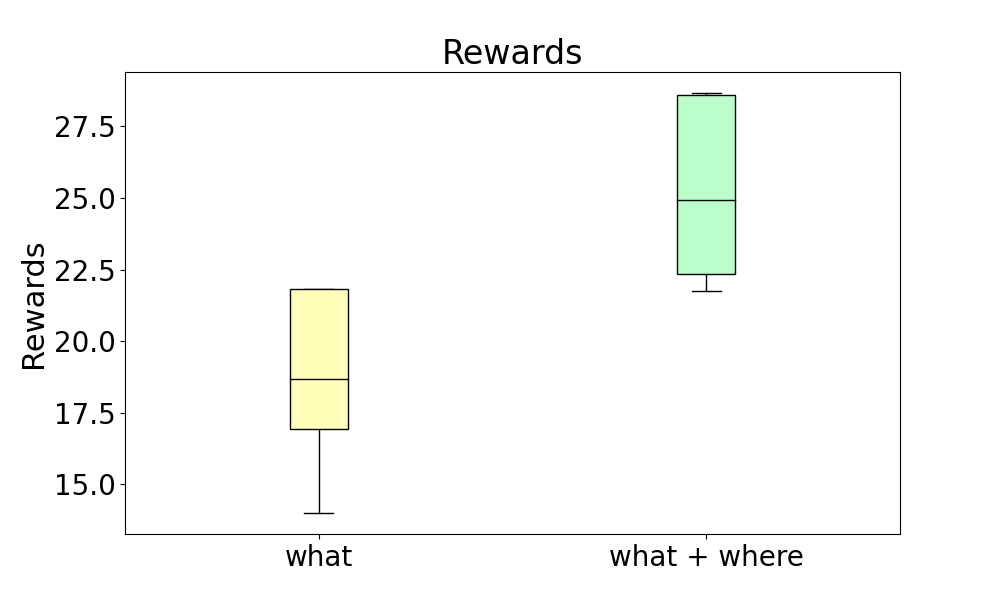}
    \caption{Success rates and mean rewards comparing ``what" and unified ``what" + ``where" representations. Each configuration was trained five times and evaluated across 100 trajectories.}
    \label{fig:what_where}
\end{figure*}

Although the initial slots encode intrinsic object properties (such as possible dynamics or affordances) distilled from temporal cues observed in the videos during pretraining, they are still produced by the invariant slot attention mechanism. As a result, they may lack crucial spatial information needed for effective reinforcement learning during action execution. We addressed this limitation in Section~\ref{sec:proposed_method} by incorporating ``where" vectors through concatenation. To evaluate their contribution, we compare imitation learning performance with and without their inclusion. The results, shown in Figure~\ref{fig:what_where}, highlight the significance of incorporating spatial information in the slot vectors.

\subsection{Results Discussion}
In our experimental evaluation, the representations produced by our encoder consistently outperformed those of other state-of-the-art encoders, both in terms of success rate and mean reward. In the imitation learning setting (Table~\ref{table:imitation_results}), it achieved a mean success rate of $0.61$ and a mean reward of $25.25$, outperforming the closest competitor (ResNet50) by $8.93\%$ and $19.39\%$ in the respective metrics. Additionally, our method exhibited the lowest variance in success rate (alongside DINOv2) and comparable variance in mean reward ---much smaller than that of the next best-performing encoder, ResNet50--- suggesting consistency across runs. Moreover, our method uses a more compact representation than the second-best model (ResNet50), and a comparable representation size to the next best-performing models (CLIP and DINOv2). Similar observations hold for the Implicit Q-Learning scenario, where our method achieved a success rate of $0.57$ ($11.76\%$ improvement over the next best), and a mean reward of $24.41$ ($19.24\%$ improvement compared to the next best). \par
%We attribute this overall performance improvement to two main factors. First, our encoder, having been trained on videos, is able to capture action-relevant temporally-infused features that result in richer scene representations, enhancing policy learning
We attribute this overall performance improvement to two main factors. First, our encoder, having been trained on videos, is able to capture temporally-infused features that result in richer scene representations, enhancing policy learning. Second, fine-tuning on videos depicting human-performed actions ---although still out-of-domain--- provides a significant advantage, as evidenced by the performance gap with the SOLV-adapted encoder, which was trained on generic out-of-domain videos. These two  conditions enabled the constructed representations to capture action-relevant object features, such as affordances and potential dynamics, thus improving the robot's visuo-motor policy decisions. %in-domain videos ---even though the demonstrations are from humans--- provides a significant advantage, as it was underscored by the performance gap with SOLV which was trained on mostly out-domain videos. 
It is also worth noting that in achieving these results, design choices discussed in Sections~\ref{sec:num_slots} and~\ref{sec:where_contr} had a tangible impact, as the chosen number of slots and the addition of the spatial components ensured, through our experimentation, the most optimal segmentation and the best task performance, respectively.

%Maybe mention it in the experimental setup
%In our approach, we aim to exploit the SOLV pretrained knowledge; however, using its as-is is not optimal for the specific tasks the robot will need to do in an exploration scenario. Therefore, we firstly fine-tune SOLV's weights, having incorporated a Multi-Layer Perceptron, which predicts the slots generated by the Temporal Binder's output, receiving as input the Spatial Binder's slot vectors from the cenrtal frame. This addition is done in an attempt to associate some action-specific information with the object representation vectors, 

%The main idea of this work was to leverage the capabilities of large pretrained encoders like DINOv2, 

%- Motivated by our goal to create object-centric representation of videos that will be suited to function as the state representation for reinforcement learning scenarios, 

\section{Conclusions}
%Using our method, robots can construct task-relevant representations without additional training, enabling an improved guidance towards the achievement of their goals, thus significantly enhancing robots' open-ended capabilities, aligning with the human cognitive development. 
In this work, we have demonstrated that image representations, derived from aggregating action-object-centric features ---produced via a slot attention mechanism, pre-trained in a self-supervised manner on an action database--- enhance the effectiveness of reinforcement and imitation learning algorithms. Using our method, robots can construct action-enhanced representations without additional training, enabling better goal-directed behavior, thus significantly improving robots' open-ended capabilities in a way that aligns with human cognitive development. Furthermore, our results highlight the potential of large-scale, directly-available action videos as valuable resources for visual encoder training.\par
In future work, we seek to explore segmentation methods that scale well to a larger number of regions, and also to evaluate our approach in real-world robotic scenarios and on more complex tasks.

% Bibliography
%\FloatBarrier
\bibliographystyle{IEEEtran} % Or another style that fits the guidelines
{\hypersetup{hidelinks} % Generated by IEEEtran.bst, version: 1.14 (2015/08/26)

 % Assumes your .bib file is named references.bib

%\printbibliography
\end{document}